\documentclass[journal]{IEEEtran}
\usepackage{cite}
\ifCLASSINFOpdf
  \usepackage[pdftex]{graphicx}
\else
  \usepackage[dvips]{graphicx}
\fi
\usepackage{amsmath}
\usepackage{array}

\usepackage{amssymb}
\usepackage[ruled]{algorithm2e}
\usepackage{algorithmic}

\begin{document}
\title{Locality-aware Channel-wise Dropout for Occluded Face Recognition}

\author{Mingjie~He,
        Jie~Zhang, ~\IEEEmembership{Member,~IEEE,}
        Shiguang~Shan,~\IEEEmembership{Senior Member,~IEEE,}
        Xiao Liu,
Zhongqin Wu,
        and~Xilin~Chen,~\IEEEmembership{Fellow,~IEEE}% <-this % stops a space
\thanks{M. He, J. Zhang, S. Shan, and X. Chen are with the Key Laboratory of Intelligent Information Processing of Chinese Academy of Sciences, Institute of Computing Technology, CAS, Beijing 100190, China, and also with the University of Chinese Academy of Sciences, Beijing 100049, China. S. Shan is also with CAS Center for Excellence in Brain Science and Intelligence Technology, Shanghai 200031, China, and with Peng Cheng Laboratory, Shenzhen 518055, China (e-mail: \{hemingjie, zhangjie, sgshan, xlchen\}@ict.ac.cn).

X. Liu and Z. Wu are with the Tomorrow Advancing Life Education Group (TAL), Beijing 100080, China (e-mail: \{liuxiao15, wuzhongqin\}@tal.com)
}% <-this % stops a space
\thanks{ Corresponding author: Shiguang Shan.}}

\maketitle

\begin{abstract}
Face recognition remains a challenging task in unconstrained scenarios, especially when faces are partially occluded. To improve the robustness against occlusion, augmenting the training images with artificial occlusions has been proved as a useful approach. However, these artificial occlusions are commonly generated by adding a black rectangle or several object templates including sunglasses, scarfs and phones, which cannot well simulate the realistic occlusions. In this paper, based on the argument that the occlusion essentially damages a group of neurons, we propose a novel and elegant occlusion-simulation method via dropping the activations of a group of neurons in some elaborately selected channel. Specifically, we first employ a spatial regularization to encourage each feature channel to respond to local and different face regions. In this way, the activations affected by an occlusion in a local region are more likely to be located in a single feature channel. Then, the locality-aware channel-wise dropout (LCD) is designed to simulate the occlusion by dropping out the entire feature channel. Furthermore, by randomly dropping out several feature channels, our method can well simulate the occlusion of larger area. The proposed LCD can encourage its succeeding layers to minimize the intra-class feature variance caused by occlusions, thus leading to improved robustness against occlusion. In addition, we design an auxiliary spatial attention module by learning a channel-wise attention vector to reweight the feature channels, which improves the contributions of non-occluded regions. Extensive experiments on various benchmarks show that the proposed method outperforms state-of-the-art methods with a remarkable improvement.
\end{abstract}

% \fi
\IEEEpeerreviewmaketitle
\section{Introduction}

With the huge success of deep learning, a remarkable improvement has been achieved for face recognition under controlled settings (i.e., occlusion-free images, near-frontal poses, neutral expressions, normal illuminations, etc.). However, in realistic unconstrained scenarios, face recognition remains a challenging task due to various factors including very large poses, very low resolution and occlusions. Among these factors, the occlusion is an intractable problem which leads to a severe degeneration in recognition accuracy.

The occlusions always bring about two primary issues, i.e., the missing of facial information and the noise from occlusion. To improve the robustness against occlusion, many efforts~\cite{cheng2015robust,zhao2017robust,pathak2016context,iizuka2017globally,li2017generative,yuan2019face} have been made to recover the occluded faces. The work in~\cite{zhao2017robust} uses a multi-scale spatial long short-term memory (LSTM) encoder to encodes occluded face patches, and then another LSTM is employed to reconstruct the occlusion-free face image. Based on the Generative Adversarial Network (GAN)~\cite{goodfellow2014generative}, another work~\cite{li2017generative} proposes a face completion model to generate visually plausible contents for the occluded face regions. Although a huge progress has been made, the performance of occlusion removal is still far from satisfactory. The main reason is that these methods are commonly trained with artificial occluded images. For instance, the images are manually generated by randomly putting a black rectangle or several object templates including sunglasses, scarfs, phones, and cups on them, which differ significantly from real-life occlusions. Therefore these methods always suffer from poor generalizations under realistic scenarios. Besides, how to well recover the occluded face regions as well as preserve the identity information is another challenge for these methods.

Another type of methods focus on suppressing the noise caused by occlusions. They attempt to discard the corrupted feature elements which are extracted from the occluded regions~\cite{min2011improving,chen2011occluded,park2015partially,wan2017occlusion,song2019occlusion,ding2020masked}. PDSN~\cite{song2019occlusion} builds a mask dictionary in advance by comparing the features extracted from normal faces and those from occluded faces. During the recognition process, the occluded facial regions are detected by a segmentation network and then the noise is removed by discarding the corrupted feature elements retrieved from the mask dictionary. To some extent, the occlusion discarding method relieves the influence of occlusions. However it may be non-trivial to precisely detect the real-world occlusions which usually have various shapes and textures since the occlusion detection modules are also trained with artificial occlusions. Even if the corrupted feature elements have been located perfectly, directly zeroing out them will incorporate a peculiar pattern into the final feature representation. Moreover, due to the uncertain size of the occluded region, the final feature will have an unfixed number of valid elements. Thus, the traditional metrics designed for the fixed-length vectors may fail for evaluations.

\begin{figure*}[t]
    \centering
    \includegraphics[width=1\linewidth]{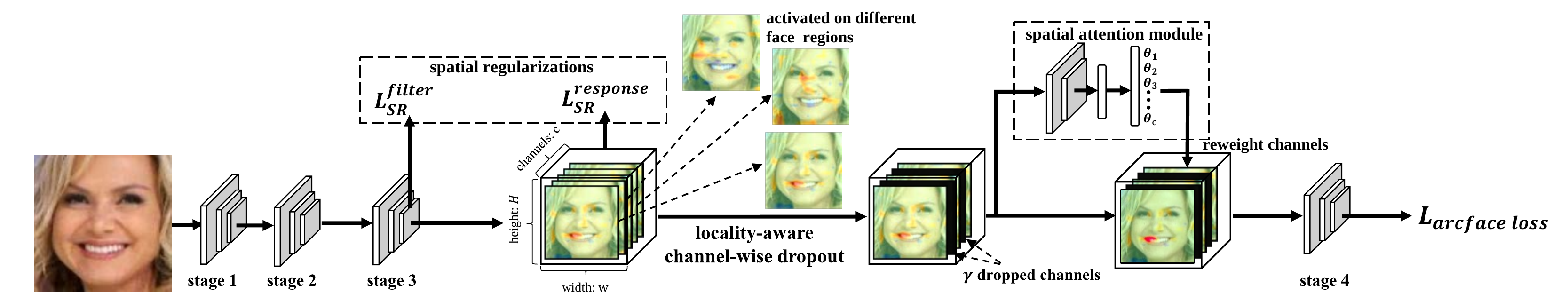}
    \vspace*{-15pt}
\caption{The overall architecture of the proposed locality-aware channel-wise dropout (LCD). Two spatial regularization losses are employed to encourage each feature channel to respond to local and different face regions. In this way, the activations affected by an occlusion in a local region are more likely to be located in a single feature channel. Then, our LCD achieves a feature-level occlusion simulation by randomly discarding a group of feature channels. Furthermore, the auxiliary spatial attention module learns a channel-wise attention vector to reweight the feature channels, which improves the contributions of non-occluded regions.
    }
    \label{figure:method}
    \vspace*{-15pt}
\end{figure*}

One simple way of tackling occluded face recognition problem is to leverage a massive number of occluded faces in the real world to directly train deep neural networks, which are forced to learn occlusion-robust face features. However, it is hard to collect such a training set. As an alternative, augmenting the training images with artificially synthesized occlusions has been studied and significant improvements have been witnessed in ~\cite{lv2017data,osherov2017increasing,trigueros2018enhancing,shaobiased}. However, the image-level occlusion simulation is still not an elegant solution since the artificial occluded images are also generated by using limited hand-craft occlusion templates, which cannot fully represent the arbitrary patterns in realistic occlusions.

The common issue of all the aforementioned methods is that the artificially synthesized occlusion cannot precisely represent the real-world occlusions, leading to poor generalizations under realistic scenarios. In this paper, we propose a novel and elegant method which can better simulate realistic occlusions. Here, we hold the opinion that the occlusion essentially damages a group of neurons. To synthesize different occlusions, a natural approach is to dropout the activations of various neurons. However, conventional dropout operation cannot simulate the real-life occlusion. The reason is that real-life occlusion affects a contiguous region in an activation map, while the conventional dropout operation discards discrete activations. To better simulate the feature damaged by occlusions, we propose the revised dropout method, namely the locality-aware channel-wise dropout (LCD) to drop a group of activations which are affected by the same facial occlusion. For conventional neural networks, partial occlusion usually affects activations across multiple feature channels, which makes it difficult to simulate occlusions with a few channels. Inspired by the~\cite{Novotny_2017_CVPR}, we encourage each feature channel to only respond to a local and different face region via the spatial regularization. As shown in Fig.~\ref{figure:method}, the heat maps of 3 different feature channels are visualized and it can been seen that these channels respond to various face regions. Therefore, the activations affected by a partial occlusion are more likely to be located in a single feature channel. Then, LCD can simulate the occlusion at the local region by dropping out the entire feature channel. It can also simulate the occlusion of larger area by randomly dropping out more feature channels.

Setting our LCD at a middle depth in the neural network can encourage its succeeding layers to minimize the intra-class feature variance caused by occlusions, leading to improved robustness against occlusion. Furthermore, we design an auxiliary spatial attention module which learns a channel-wise attention vector to reweight the channels during the feature extraction process. After jointly trained with our LCD, the deep network is optimized to focus more on the channels which are related to the non-occluded regions and suppress others which are affected by the occluded regions.

Compared with previous works, our method has three major advantages: 1) our method does not require artificially synthesized occlusions. Instead, it gracefully simulates the realistic occlusions in intermediate features. 2) Different from previous works which use additional module to detect or recover the occluded region, our method imposes minor increase in model complexity during the inference phase. 3) Our method is a more practical approach which can be seamlessly integrated with any existing face recognition method for improving robustness to occlusions.

The main contributions of this paper are summarized as follows:

\begin{itemize}
\item We propose a novel method to better simulate realistic occlusions by dropping a group of activations in intermediate features. It significantly improves the robustness to occlusions by encouraging the neural network to emphasize on learning discriminative features from the non-occluded face regions.
\item An auxiliary spatial attention module is designed to improve the contributions of non-occluded regions by adaptively reweighting the feature channels.
\item Our method significantly outperforms the state-of-the-art methods on IJB-C, LFW and MegaFace benchmarks, especially on the IJB-C dataset with large-scale real-occluded face images.
\end{itemize}

The rest of the paper is organized as follows. Section~\ref{sec:related} gives a brief overview of the related works. Section~\ref{sec:method} introduces the detailed formulation of the proposed method, followed by a discussion with other state-of-the-art methods in Section~\ref{sec:discussion}. Section~\ref{sec:experiment} presents the ablation study and the experimental results on three databases. Finally, the conclusion is summarized in Section~\ref{sec:conclusion}.

\section{Related Works}
\label{sec:related}
The existing occlusion robust face recognition methods can be broadly grouped into the following three categories: occluded face completion methods, occlusion-aware discarding methods and occlusion-robust feature extraction methods. In this section, we provide a brief overview of the recent works which are most relevant to this paper.

\subsection{Occluded Face Completion Methods}
The occluded face completion methods are pixel-level approaches which aim to recover the occluded face regions. Considering the low-rank property of non-occluded images, early works attempt to solve the problem by using robust principal component analysis (PCA) to reconstruct the corrupted low-rank face images~\cite{park2005glasses,fidler2006combining,candes2011robust,liu2012robust,zhang2015double}. The work in~\cite{candes2011robust} proposes the robust principal component analysis (RPCA) which improves the performance of removing shadows from face images. In~\cite{liu2012robust}, an important extension of RPCA, namely the low-rank representation (LRR) is presented to extend the recovery of corrupted images from a single subspace to multiple subspaces. Both RPCA and LRR assume that the occluded pixels are sparse, while real-world face images usually contain dense occlusions, which make the matrix non-sparse. To this end,~\cite{zhang2015double} presents the double nuclear norm-based matrix decomposition to remove the dense occlusions.

Recently, more works resort to the deep learning for improving the occluded face completion~\cite{cheng2015robust,zhao2017robust,pathak2016context,iizuka2017globally,li2017generative,yuan2019face,qian2019unsupervised}. In~\cite{cheng2015robust}, the difference between the activation values of two stacked sparse denoising auto-encoders (SSDAs) are used to indicate occluded and un-occluded face regions. Then, the final occlusion-free image is reconstructed by transferring the encoding activations of the un-occluded region to the occluded region. In~\cite{pathak2016context}, the context encoder combines the auto-encoder architecture with context information of the occluded part to produce visually pleasing images. To further improve the context encoder,~\cite{iizuka2017globally} introduces both global and local discriminators. Specifically, the global discriminator pursues the global consistency of the overall image and the local discriminator looks at a small area centered at the reconstructed region to judge the quality in details. To ensure the new generated contents more photo-realistic, a semantic parsing loss is developed in~\cite{li2017generative}. To refine local face textures, a 3D morphable model (3DMM) is utilized in~\cite{yuan2019face} to further assist the learning of the local discriminator. In the unsupervised face normalization method (FNM) ~\cite{qian2019unsupervised}, multiple local discriminators are integrated into a novel unsupervised framework. It generates impressive high quality faces which have dispelled various face variations including occlusions.

Overall, the above-mentioned face completion methods have shown promising results of transforming occluded faces to un-occluded ones. However, two major issues still remain. Firstly, except for a few methods (e.g., the FNM~\cite{qian2019unsupervised}), most previous methods require pair-wise training data (i.e., one occluded face and one un-occluded face of the same person). Unfortunately, such training sets containing a mass of pair-wise faces with natural occlusion are extremely rare. An alternative and commonly used approach is using synthetically occluded faces. However, these synthetically occluded faces, e.g., using manually designed occlusion templates or a black/white rectangle can not fully represent natural occluded faces. Secondly, the faces generated by GAN-based methods are usually visually pleasing. However, how to remove the occlusion while preserving the identity information still remains a challenging problem.

\subsection{Occlusion-Aware Discarding Methods}
These approaches aims to remove the noise caused by occlusions with two pipelines, which either discards the occluded pixels before the face feature extraction, or discards the corrupted feature elements during the feature extraction.

Following the former pipeline, some works detect the occlusions first and then extract a feature representation from the non-occluded regions only. The early works~\cite{oh2008occlusion,min2011improving,chen2011occluded,park2015partially} usually employ a nearest neighbor classifier (NNC) or a support vector machine (SVM) to classify the occluded face regions. Since these methods are designed based on the traditional feature descriptors, it is non-trivial for them to obtain discriminative ability for face recognition in complex scenarios. Recently, the LPD~\cite{ding2020masked} designs a neural network to locate the latent facial parts which are less affected by a specific occlusion (i.e., the respirator), and then extracts discriminative features from the selected latent part.

Following the spirit of the latter pipeline, the mask leaning methods~\cite{wan2017occlusion,song2019occlusion} locate and discard the corrupted feature elements rather than the occluded pixels. In~\cite{wan2017occlusion}, the MaskNet adaptively learns feature masks for occluded face images and automatically assigns lower weight to the hidden units activated by the occluded face regions. The PDSN~\cite{song2019occlusion} establishes a mask dictionary to represent the correspondence between the occluded facial block and the corrupted feature elements. During the testing phase, a segmentation network is employed to detect the occluded facial blocks, and then the corrupted feature elements are set to zero by retrieving the relevant dictionary items.

Although above-mentioned occlusion-aware discarding methods can alleviate the occlusion issue, completely discarding the occluded regions still takes the risk of reducing the system reliability. Firstly, precise and fine-grain occlusion detection is an essential prerequisite for these methods. However, such occlusion detection is non-trivial to obtain and a coarse detection will increase the risk of losing discriminative information or inducing unreliable information. Furthermore, even if the occluded face regions or the corrupted feature elements have been located perfectly, directly zeroing out the corrupted elements still incorporates a peculiar pattern into the final feature, which may harm the recognition accuracy. Moreover, due to the arbitrary occlusion, the feature vectors for occluded faces have unfixed number of valid elements. Thus, the conventional metrics for fixed-length vectors are not fully applicable. Although works on the instance-to-class distance~\cite{hu2013robust} and the reconstruction-based similarity measurement~\cite{he2018dynamic} are proposed to tackle the above problem, the commonly used window sliding makes them relatively time-consuming. On the other side, directly ignoring occluded elements will potentially break the global cues of face images such as chin contours, which is also harmful to face recognition system.

\subsection{Occlusion-Robust Feature Extraction Methods}
Directly learning occlusion-robust feature is the most straightforward and effective way to handle occlusion face recognition. For this purpose, many efforts are devoted to seeking a feature space that is less affected by occlusion and meanwhile preserves the discriminative capability of distinguishing identities. Many works seek such feature space via the sparse representation classification (SRC)~\cite{wright2008robust}, in which the occluded image is represented by a linear combination of training samples plus a sparse constraint term accounting for occlusions. The LR-LUM~\cite{dong2019low} combines both the robust sparsity constraint and the low-rank constraint which outperforms previous methods in handing structured occlusion such as sunglasses and scarfs. Other works~\cite{yang2013gabor,jia2012robust,yang2017joint} extend the sparse representation by combining more discriminative feature descriptors. In~\cite{yang2017joint}, the JCR-ACF proposes a joint and collaborative representation with local adaptive convolution feature, which can improve the recognition accuracy by employing information from different local face regions. Although these SRC-based methods have made considerable progresses, they do not generalize well in practical scenarios since they requires that the test samples have identical identities with a pre-defined close-set.

Owing to leveraging massive training sets, recent deep learning methods reveal significant superiority on face recognition. Enlarging the training datasets with sufficient occluded faces may be an effective way to improve the occlusion robustness of face embedding. Unfortunately, such training sets of a mass of identities are extremely rare. As an alternative solution, synthetic image augmentation method has been studied by previous works~\cite{lv2017data,osherov2017increasing,trigueros2018enhancing,shaobiased}. The work in~\cite{lv2017data} enriches the training set by synthesizing occluded faces with various pre-defined hairstyles and glasses templates. Although an accuracy gain has been witnessed, the diversity of the manually designed occlusion templates needs to be further improved. More recently, BFL~\cite{shaobiased} proposes an enhanced augmentation schema to randomly generate multi-scale spatially occluded samples and then modifies the loss to balance the impact of normal and occluded samples for training. To some extent, these augmentation methods improve the robustness of neural network, but the discrepancy between the synthetic and real occluded faces still limits the further improvements of the robustness.

Another line of researches~\cite{li2018harmonious,wang2020region,yin2019towards} focus on  the attention mechanism for robust feature extraction. The state-of-the-art method named InterpretFR~\cite{yin2019towards} employs a Siamese network to compare the feature elements from a normal face with and without synthetic occlusion. Then it encourages the neural network to identify the input face solely based on the feature elements which are less sensitive to occlusions. However, it still need synthesizing occluded faces by using artificial occlusion templates and fails to generalize well on unseen occlusions. In contrast, our method can realistically simulate arbitrary occlusions via dropping out a random group of filter responses, leading to an improved performance under real world scenarios.

\section{METHODOLOGY}
\label{sec:method}
\subsection{Overview}
The proposed method attempts to learn occlusion robust face features by simulating occlusions during the training process. Different from previous methods which augment face images with synthesized occlusions, we propose to directly simulate the influence of arbitrary occlusions on intermediate features. As the occlusion essentially damages a group of neurons, we propose a revised dropout method, namely the locality-aware channel-wise dropout (LCD) to simulate occlusions by dropping a group of feature channels. Considering that, for the conventional neural networks, the occlusion usually affects the activations from most channels which is even impossible to simulate the occlusions by dropping several channels. Therefore, we first employ a spatial regularization to encourage each feature channel to respond to different face regions. In this way, the activations affected by the partial occlusion are more likely to be located in a single feature channel. Then, our LCD can simulate the occlusion by dropping out the entire feature channel, in which sense we name this method as locality-aware channel-wise dropout. Moreover, to improve the contributions of non-occluded regions for learning occlusion robust features, we design an auxiliary spatial attention module to reweight the feature channels. The whole framework shown in Fig.~\ref{figure:method} is end-to-end trainable and can be easily applied on any existing convolution neural network.

\subsection{ Locality-aware Channel-wise Dropout}
\subsubsection{\textbf{Spatial regularization of feature channels}} We employ the spatial regularization~\cite{Novotny_2017_CVPR} to encourage each feature channel to respond to local and different face regions. The spatial regularization consists of two loss functions. The first one is a filter orthogonal loss:
\begin{equation}
\label{filter_regular}
L_{{SR}}^{filter} = \sum\limits_{i \ne j} {\left| {\sum\limits_p {\frac{{\left\langle {\mathbf{w}_i^p,\mathbf{w}_j^p} \right\rangle }}{{{{\left\| {\mathbf{w}_i^p} \right\|}_F}{{\left\| {\mathbf{w}_j^p} \right\|}_F}}}} } \right|} ,
\end{equation}
where $\mathbf{w}_i^p$ denotes the $p$-th column of convolution filter $\mathbf{w}_i$. The Eq.~\ref{filter_regular} encourages the orthogonality of filters, which make the different filters more likely to respond to various face regions. Besides, the other loss function is employed to further enhance such characteristic by directly penalizing the correlations between the filters' responses:
\begin{equation}
\label{respone_regular}
L_{{SR}}^{response} = \sum\limits_{i \ne j} {\left\| {\frac{{\left\langle {{\mathbf{f}_i},{\mathbf{f}_j}} \right\rangle }}{{{{\left\| {{\mathbf{f}_i}} \right\|}_F}{{\left\| {{\mathbf{f}_j}} \right\|}_F}}}} \right\|} ^2 ,
\end{equation}
where $\mathbf{f}_i$ denotes the response of $i$-th filter (i.e., the $i$-th channel of features).

\subsubsection{\textbf{Simulating occlusions via Channel-wise Dropout}}

Given an input feature map $\textbf{F} \in \mathbb{R}{^{c \times h \times w}}$, we first generate an all-one mask matrix $\textbf{M} \in \mathbb{R}{^{c \times h \times w}}$ with the same size as $F$, where $c$, $h$, $w$ denotes the channel number, the height, the width of the feature map,  respectively. Second, we randomly sample $\gamma$ distinct channel indexes $\{r_1,r_2,..,r_{\gamma}\}$ from the $c$ channels. Then, the mask values for these channels are set to zero:

\begin{equation}
\label{fs-mask}
M_{i,j,k}=
\begin{cases}
0& i \in \{r_1,r_2,..,r_{\gamma}\} \\
1& \text{others}
\end{cases}.
\end{equation}

Finally, the output of the LCD is obtained by the product of the mask matrix and the input:
\begin{equation}
\label{fs-mask}
\textbf{F}_{drop} = \textbf{F} \circ \textbf{M},
\end{equation}
where $\circ $ denotes Hadamard product. As shown in Fig.~\ref{figure:method}, all the $\gamma \times h \times w$ feature elements within the $\gamma$ feature maps will be dropped out to zero. At the training stage, our LCD performs as an effective occlusion simulation and encourages the network to identify the input face solely based on the remaining feature, make it more robust to occlusions. It should be mentioned that similar to the conventional dropout, the LCD is not employed during the inference process.

In our method, $\gamma$ is a crucial parameter which controls how many feature channels will be dropped out. In other words, it determines how many face regions will be discarded during the training process. A larger $\gamma$ simulates a more severe occlusion where more regions in the face are occluded. To simulate the complex pattern of realistic occlusions, a dynamic $\gamma$ for each training samples is required. With this in mind, the $\gamma$ is designed as a stochastic variable satisfying uniform distribution within $\left[ {{\gamma _{\min }},{\gamma _{\max }}} \right]$. A larger $\gamma _{\max }$ (e.g., $\gamma _{\max }=0.6*c$) is recommended to improve the robustness of severer occlusions.

Since the activations of convolutions layers are commonly normalized by batch-normalization (BN) layers, how to make the LCD compatible with the BN layers is worth exploring. In a conventional BN layer, the feature elements sharing the same channel index will be normalized together. However, this process will have some problems when applying the LCD before the BN layer. Specifically, for a mini-batch with $n$ samples, let $x_{t,i,j,k}$ denote its $t,i,j,k$-th element in the $n \times c \times h \times w$ feature tensor. The conventional BN layer computes the mean for $i$-th channel:

\begin{equation}
\label{BNmean}
u_i=\frac{1}{nhw}\sum_t^n \sum_j^h \sum_k^w x_{t,i,j,k}.
\end{equation}

However, since features of a sub-set channels are set to zero, the number of valid samples for the $i$-th channel is no longer equal to $n$. To resolve this problem, the calculation of $u_i$ must be modified to:

\begin{equation}
\label{fs-mean_modified}
u_i=\frac{1}{(n-\eta_i)hw}\sum_t^n \sum_j^h \sum_k^w x_{t,i,j,k},
\end{equation}
where $\eta_i$ denotes the number of training samples which have zero values in the $i$-th channel. Besides, the calculation of the channel variance also requires similar modification. To be free from these modifications, always setting the LCD after the conventional BN layer is an alternative solution. Moreover, this simple setting is even more favorable to obtain stable BN parameters as all the training sample are involved in the computation of mean and variance.

\subsection{Spatial Attention Module}

\begin{figure}[t]
    \centering
    \includegraphics[width=7cm]{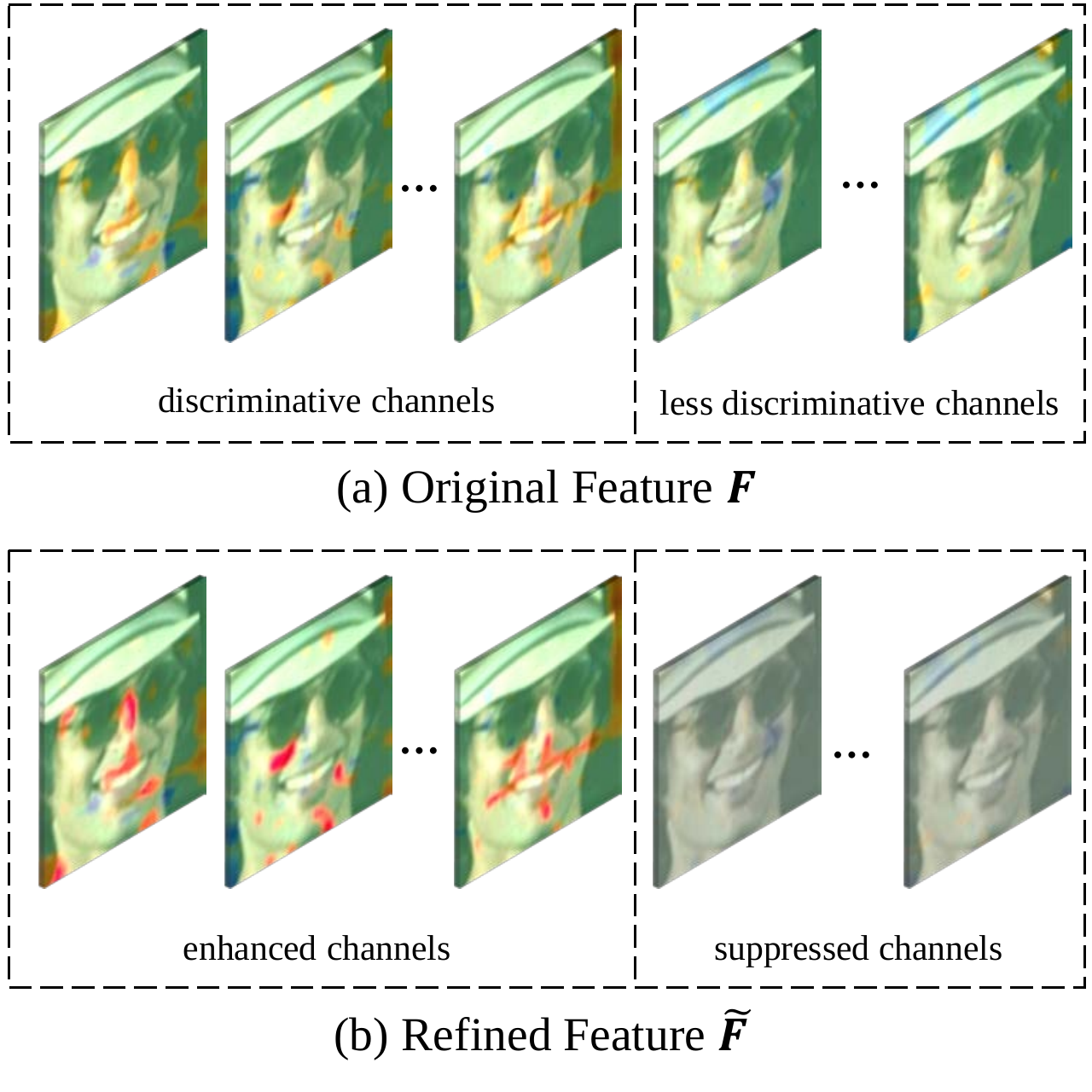}
    \vspace*{-15pt}
\caption{(a) Due to the occlusion by sunglasses and hat, only the feature channels related to the non-occluded regions contain discriminative information, while other channels are corrupted by occlusions. (b) The spatial attention module attempts to reweight the features, making the network focus on the channels which are related to the non-occluded regions.
     }
    \label{figure:attention}
    \vspace*{-15pt}
\end{figure}
In an occluded face image, the feature channels responded to the occluded regions will not contain discriminative information for face recognition. Besides, they potentially contains the noise from the occluded regions. To be free from this, we design an attention module to reweight the feature channels, making the network focus more on the feature channels which are related to the non-occluded regions. It is worth noting that, by utilizing the aforementioned spatial regularization, each feature channel is encouraged to respond to local and different face regions. In this sense, reweighting these feature channels actually performs as a spatial-wise attention approach, which is named as the spatial attention module (SAM).

For the input feature map $\mathbf{F}=[\mathbf{f}_1,\mathbf{f}_2,...,\mathbf{f}_c]$, our goal is to learn an attention vector $\boldsymbol{\theta}=[\theta_1,\mathbf{\theta}_2,...,\mathbf{\theta}_c]$ which controls the weight of each feature channel. As shown in Fig.~\ref{figure:method}, a light-weight module is designed to learn the attention vector $\boldsymbol{\theta}$. Specifically, we first employ a $1 \times 1$ convolution layer and a fully-connected layer to obtain a global view of all input feature maps. Then, another fully-connected layer is utilized to extract the channel-wise attention vector. It is worth noting that previous attention block in SENet~\cite{hu2018squeeze} utilizes a global average pooling as a global descriptor, while in our method the global pooling of dropped zero value feature maps will suffer from local minimum. To this end, the $1 \times 1$ convolution layer in our proposed module is a necessary and effective information aggregation strategy to achieve attentions.

The refined feature map $\widetilde{\mathbf{F}}$ is obtained by channel-wise multiplication between each feature map $\mathbf{F}_x$ and its corresponding attention scalar $\mathbf{\theta}_x$:
\begin{equation}
\label{attention}
\widetilde{\mathbf{F}}=\mathbf{F}\cdot\boldsymbol{\theta}=[\mathbf{f}_1\cdot\theta_1,\mathbf{f}_2\cdot\theta_2,...,\mathbf{f}_c\cdot\theta_c].
\end{equation}

Formally, the Algorithm~\ref{alg:tn} summarizes the training process using the locality-aware channel-wise dropout and the spatial attention module.

\begin{algorithm}%[H]
\label{alg:tn}
\SetAlgoNoLine
\caption{Training with the locality-aware channel-wise dropout and the spatial attention}
\KwIn{ a mini-batch with $n$ training images and their labels, $\gamma _{\min }$, $\gamma _{\max }$, the loss weights ${\alpha}$ and ${\beta}$.}

\While{not converged}{
    Forward propagation to get the intermediate features: \\
     \centerline{$\mathbf{F}_i \in \mathbb{R}{^{c \times h \times w}}, i\in [1,n]$; }
    Compute the filter orthogonal loss $L_{{SR}}^{filter}$\;
    Compute the response orthogonal loss $L_{{SR}}^{response}$\;
    \For{each $i\in [1,n]$} {
        Randomly select the $\gamma_i$ :\\
         \centerline{$\gamma_i \sim Uniform(\gamma _{\min},\gamma _{\max})$;}
        Randomly sample $\gamma_i$ channels from the $\mathbf{F}_i$\;
        Set all the values in these $\gamma_i$ channels to zero\;
        Compute the attention vector $\boldsymbol{\theta}_i$\;
        Compute the refined feature: \\ \centerline{$\widetilde{\mathbf{F}_i}=\mathbf{F}_i\cdot\boldsymbol{\theta}_i$}

    }
    Compute the output of the succeeding layers\;
    Compute the softmax loss ${L_{arcface}}$\;
        Compute the total loss $L_{total}$:\\
        \centerline{
        $L_{total}=L_{arcface}+{\alpha}L_{{SR}}^{filter}+{\beta}L_{{SR}}^{response}$;
        }
    Backward propagation\;
    Update the weights of the neural network\;

}

\KwOut{the trained neural network.}
\end{algorithm}

\section{Discussion}
The proposed LCD method can be categorized as a form of structured dropout. Conventionally, other structured dropout methods are designed for the purposes of alleviating the over-fitting issue and improving the generalization ability. In contrast, our LCD is designed for simulating facial occlusions and achieving occlusion robustness. In this section, we discuss the comparisons of our LCD with two representative structured dropout methods, i.e., the DropBlock~\cite{ghiasi2018dropblock} and the weighted channel dropout (WCD)~\cite{hou2019weighted}. Furthermore, we also provide a comparison with the occluded face recognition method InterpretFR~\cite{yin2019towards} which is most relevant to our method.
\label{sec:discussion}
\subsection{Differences from DropBlock~\cite{ghiasi2018dropblock}}
Both our method and DropBlock drop some activations in the intermediate feature layers, but they differ in two aspects. 1) The Dropblock drops partial regions within each feature channel for enhancing feature learning while our method employs spatial regularization to enforce each feature channel to respond to local face regions and further drop several feature channels to simulate partial occlusions. 2) Our method further proposes spatial attention module to improve the contributions of undamaged neurons to further promote the robustness to occlusion. Experimental results on IJB-C datasets show our method significantly outperforms Dropblock for occluded face recognition.

\subsection{Differences from the Weighted Channel Dropout~\cite{hou2019weighted}}
Both our method and the weighted channel dropout (WCD) drop channels but for different purposes and in different ways. Firstly, in terms of designing purpose, the WCD is for alleviating the over-fitting issue when fine-tuning neural network on small datasets, while our method aims to improve the robustness to partial occlusion for face recognition. Secondly, in terms of what channels to drop, the WCD drops out the feature channels which have relatively lower activation magnitudes, while in our method each feature channel is enforced to respond to features corresponding to some local facial region and thus can be dropped to simulate the feature corruption due to the occlusion of that region. Comparison experiments on IJB-C datasets show the superiority of our method in handling occluded face recognition problem.

\subsection{Differences from InterpretFR~\cite{yin2019towards}}
Both our method and InterpretFR resort to occlusion robust feature learning for tackling face recognition under occlusions. However, we have two major differences. 1) InterpretFR synthesizes occlusions by putting black boxes on faces, which is unrealistic and the performance may severely degenerate under real world scenarios. Instead of simply utilizing fixed templates to synthesize occlusion, we propose the locality-aware channel-wise dropout to simulate more realistic occlusions, leading to performance improvement for occluded face recognition. 2) During training, InterpretFR masks out the elements of the final face representation sensitive to the occlusion, which may be sub-optimal as it only leverage remaining features for recognition without considering information recovering. Differently, we perform channel-wise dropout on the stage 3 of ResNet and design spatial attention module in the stage 4 to implicitly recover the absent information as verified by experiment in Section~\ref{sec:stage4}, which is more favorable to tackle face recognition under occlusions.

\section{Experiments}
\label{sec:experiment}
\subsection{Datasets}
The training images are collected from the MS-Celeb-1M dataset~\cite{guo2016msceleb}. Since this dataset contains many labeling noise, we manually clean it and finally collect 3.7 Million images from 50K identities. The revised dataset is used as our training set for the experiments. We extensively test our method on three popular benchmarks, including IJB-C~\cite{maze2018iarpa}, MegaFace~\cite{Kemelmacher-Shlizerman_2016_CVPR} and LFW~\cite{LFWTech}.

\textbf{IJB-C}. As an extension of previous IJB-A~\cite{klare2015pushing} and IJB-B~\cite{whitelam2017iarpa}, the IJB-C~\cite{maze2018iarpa} is a large-scale dataset which contains 117, 542 video frames and 3, 134 images. As 57\% of the face images in IJB-C are natural occluded, this dataset is a commonly used benchmark for occlusion-robust face recognition. In this work, we employ two evaluation settings. First, we conduct comparisons on the holistic IJB-C dataset (including both occluded faces and non-occluded faces) for general evaluations. Second, to further verify the effectiveness of tackling occlusions, the occlusion subset of IJB-C (the occluded face images only) is employed for evaluations. All the two settings follow the standard IJB-C testing protocol. The true accept rate (TAR) and the false accept rate (FAR) are used as the evaluation metrics.

\textbf{MegaFace}. The MegaFace challenge 1 (MF1) benchmark~\cite{Kemelmacher-Shlizerman_2016_CVPR} evaluates how the face recognition method performs with a huge scale of distracters. Specifically, the gallery set in MF1 contains one million face distractors, and the probe set Facescrub~\cite{7025068} contains 106,863 face images of 530 identities. In the testing pipeline, each Facescrub image will be added into the galley set and the remaining images of the same identity are exploited as probes. The rank-1 identification accuracy is used as the measurement of the face recognition performance. It should be noted that the un-cleaned version of MegaFace datasets are employed in the evaluation for the fair comparison with the state-of-the-art methods.

\textbf{LFW}. The LFW~\cite{LFWTech} is a well-known unconstrained face verification benchmark. It contains 13,233 images form 5,749 identities. In our work, we follow the standard 10-fold cross validation protocol to report the mean accuracy on the 6,000 testing image pairs.
\subsection{Implementation Details}
In our experiments, the face images are aligned based on five facial landmarks detected by~\cite{7961742} and normalized to a size of 112$\times$112. We employ the ResNet-50~\cite{10.1007/978-3-319-46493-0_38} as the baseline network and implement our method on the Tensorflow~\cite{199317} platform. The Arcface loss~\cite{deng2019arcface} with the margin of 0.5 and the scale of 64 is utilized as the identification loss in our experiments. All the models in our experiments are trained on four NVIDIA TITAN XP GPUs by using SGD. The loss weight of the ArcFace loss is set to 1 and the loss weights of the filter orthogonal regularization and the response orthogonal regularization are 100 and 1, respectively.

\subsection{Where to deploy the Channel-wise Dropout}
\label{sec:stage4}
We firstly investigate the best stage to deploy the channel-wise dropout. By integrating the channel-wise dropout into three different stages of the plain ResNet-50, respectively, we construct experiments under three settings. Specifically, the channel-wise dropout is conducted on the outputs of the last $3 \times 3$ convolution layer in the stage specific to each setting.

\begin{table}[]
\caption{Performance on IJB-C occlusion subset with channel-wise applied at three different depths.}
\label{tab:depth}	
\centering
\begin{tabular}{lp{1.3cm}<{\centering}p{1.2cm}<{\centering}p{1.2cm}<{\centering}}
\hline
\multicolumn{1}{c}{Methods}             & @FAR=.0001 & @FAR=.001 & @FAR=.01 \\ \hline
Baseline                      &  39.69     & 88.52     &   95.14      \\
channel-wise dropout (stage 2)          &  53.57     & 90.55     & 95.73         \\
channel-wise dropout (stage 3)          &  57.88     & 90.80     & 95.48         \\
channel-wise dropout (stage 4)          &  37.59    & 88.96 & 95.48         \\ \hline\end{tabular}
\end{table}

Table~\ref{tab:depth} summarizes the results on the IJB-C occlusion subset of the three different settings. As seen, the channel-wise dropout in the stage 3 outperforms the baseline by an improvement up to 18.19\% in terms of TAR when FAR = 0.0001. This setting works better than the channel-wise drop conducted in the shallow stage 2. The reason behind it is that features in stage 3 have larger receptive field than those in stage 2, which can well characterize facial components, leading to better simulation of partial occlusions on faces. Besides, conducting channel-wise dropout in the stage 4 performs severely worse than that in the stage 3, or even worse than the baseline model. We argue that employing several succeeding neural layers to compensate the damaged activations plays an important role for face recognition under occlusions. We conduct a further experiment to verify this analysis as bellow.

\begin{figure}[t]
    \centering
    \includegraphics[width=9cm]{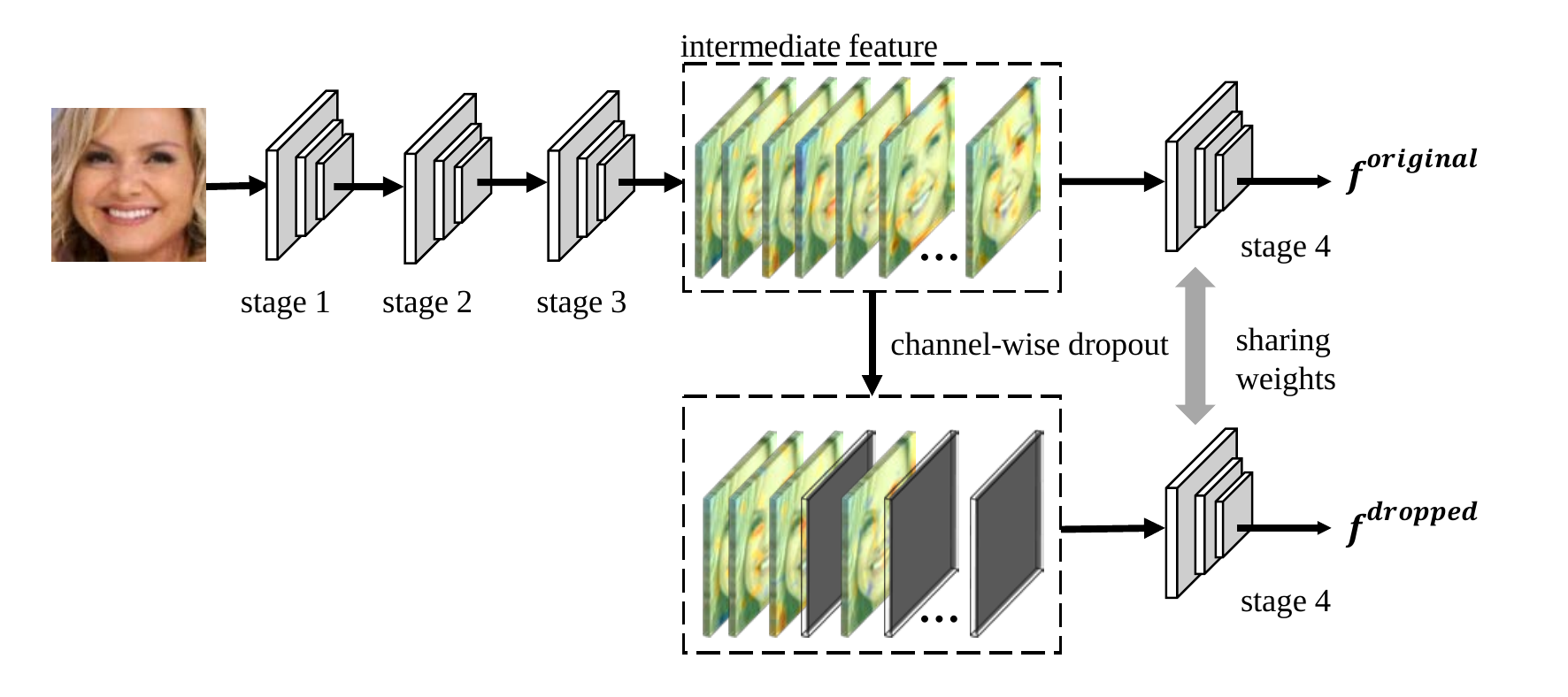}
    \vspace*{-15pt}
    \caption{The experiment designed to explore why the channel-wise dropout can improve the robustness against occlusions.}
    \label{figure:stage3}
    %\vspace*{-15pt}
\end{figure}

\begin{figure}[t]
    \centering
    \includegraphics[width=8cm]{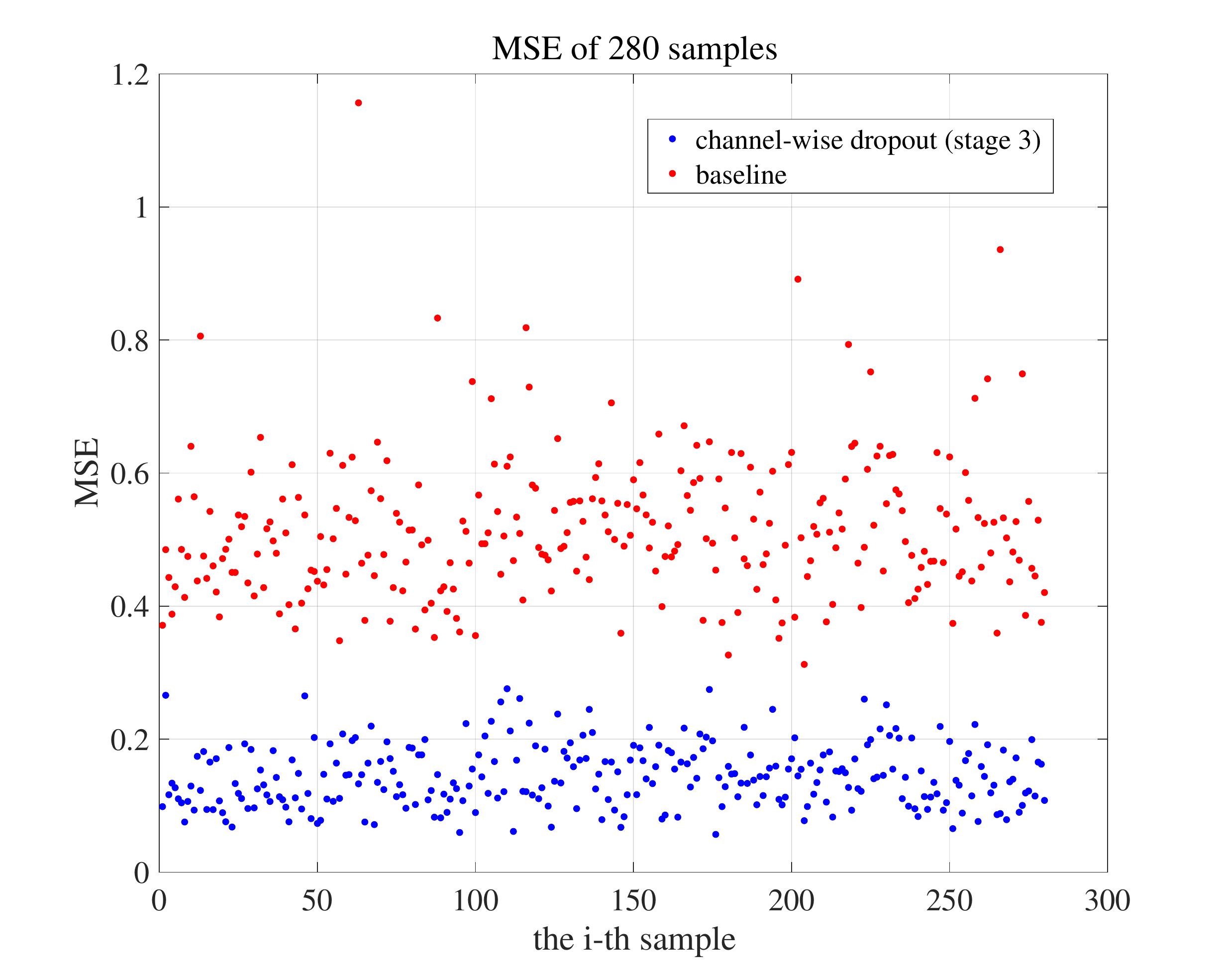}
    \vspace*{-15pt}
    \caption{The experiment designed to verify the roles of the channel-wise dropout and the neural layers behind it.     }
    \label{figure:stage3_MSE}
    %\vspace*{-15pt}
\end{figure}

For a face recognition model, we compare the similarity of output features from stage 4 with and without conducting channel-wise dropout in stage 3, as shown in Fig.~\ref{figure:stage3}. The features similarity is evaluated by mean square error (MSE) as shown in Eq.~\ref{MSE}.

\begin{equation}
\label{MSE}
MSE = \frac{1}{L}\sum\limits_{i = 1}^L {{{\left( {f_i^{original} - f_i^{dropped}} \right)}^2}},
\end{equation}
where the $L$ denotes the length of the final feature vector and $f_i$ denotes the $i$-th elements within the feature vector.

We compare the baseline mode and our model trained with the channel-wise dropout integrated in the stage 3. The results are shown in Fig.~\ref{figure:stage3_MSE}. As seen, our model trained with channel-wise dropout integrated in the stage 3 achieves notably lower MSE than baseline, which means the final features of our method on ``occluded face'' are much closer to those on clean face, leading to occlusion-robust face recognition model. The result proves that after integrating the channel-wise dropout in the stage 3, the succeeding layers in the following stage 4 play a crucial role in compensating the absent of facial information caused by occlusions. This experimental results also explain why conducting the channel-wise dropout in the stage 4 leads to poor results in Table~\ref{tab:depth}. If the occlusion is simulated in the last stage, no succeeding layers are enforced to learn features robust to occlusions. Based on this analysis, the channel-wise dropout integrated into the stage 3 of the ResNet-50 are treated as the best setting for all the following experiments.

\subsection{Ablation study}

We conduct ablation studies to evaluate the effectiveness of each component in our method. The results on the IJB-C occlusion subset are shown in Table~\ref{tab:ablation_study}. As seen, all the three components are beneficial to the performance improvement. Specifically, when only the channel-wise dropout is conducted, the TAR when FAR= 0.0001 is improved by 18.19\%. Moreover, by jointly training with the spatial regularization, a further improvement of 6.83\% is witnessed. Furthermore, by employing all the three components, our method achieves a significant improvement up to 36.58\% in terms of TAR when FAR = 0.0001, which demonstrates the effectiveness of each component of our method. For simplicity, our method consisting of all the three components is abbreviated as LCD in the following experiments.

\begin{table}[]
\caption{ The ablation study for each component of our method. \textbf{CD}: the channel-wise dropout. \textbf{SR}: the two spatial regularization losses. \textbf{SAM}: the spatial attention module.}
\label{tab:ablation_study}	
\centering
\begin{tabular}{cccccc}
\hline
baseline & CD &SR& SAM      & @FAR=.0001 & @FAR=.001 \\ \hline
\checkmark& & &                            &  39.69     & 88.52       \\
\checkmark &\checkmark & &                           &  57.88     & 90.80       \\
\checkmark &\checkmark &\checkmark &             &  64.71     & 91.50        \\
\checkmark &\checkmark &\checkmark &\checkmark  & 76.27      & 91.83   \\ \hline\end{tabular}
\end{table}

\begin{table}[t]
\caption{Performance on IJB-C dataset.}
\label{tab:IJBC_tatal}	
\centering
\begin{tabular}{lp{1.3cm}<{\centering}p{1.2cm}<{\centering}p{1.2cm}<{\centering}}
\hline
\multicolumn{1}{c}{Methods}         & @FAR=.0001 & @FAR=.001 & @FAR=.01 \\ \hline
DR-GAN~\cite{tran2018representation}                               &  -     & 73.6  & 88.2\\
PFE~\cite{shi2019probabilistic}&-&89.6&93.3\\
DUL~\cite{chang2020data}&-&90.2&94.2\\
CASIA-Net InterpretFR~\cite{yin2019towards}             &  -     & 89.2 &75.6\\
ResNet-50 InterpretFR~\cite{yin2019towards}              &  -     & 93.2 & 95.8  \\
Cutout~\cite{devries2017improved}                      & 74.7 & 94.6 & 97.7 \\
DropBlock~\cite{ghiasi2018dropblock}                   &77.3 & 94.5 & 97.6   \\
WCD~\cite{hou2019weighted}   &87.5 & 93.3& 96.1\\
\hline
LCD (ours)  & 89.8 & 95.4 & 97.5\\ \hline\end{tabular}
\end{table}

\subsection{Comparisons with state-of-the-art Methods}
\subsubsection{Evaluations on the IJB-C benchmark}

\begin{table}[t]
\caption{Performance on IJB-C occlusion subset.}
\label{tab:module}	
\centering
\begin{tabular}{lp{1.3cm}<{\centering}p{1.2cm}<{\centering}p{1.2cm}<{\centering}}
\hline
\multicolumn{1}{c}{Methods}         & @FAR=.0001 & @FAR=.001 & @FAR=.01 \\ \hline
DR-GAN~\cite{tran2018representation}                               &  -     & 66.1     & 82.4 \\
CASIA-Net InterpretFR~\cite{yin2019towards}             &  -     & 69.3     & 83.8 \\
ResNet-50 InterpretFR~\cite{yin2019towards}              &  -     & 89.8     & 93.4 \\
Cutout~\cite{devries2017improved}                      &  47.9     & 90.5     & 96.0       \\
DropBlock~\cite{ghiasi2018dropblock}                   &  51.0 &   90.4 & 95.7  \\
WCD~\cite{hou2019weighted}   &74.6 & 88.5& 93.1\\
\hline
LCD (ours)  & 76.3      & 91.8     & 95.4   \\ \hline\end{tabular}
\end{table}

We firstly evaluate our LCD on the IJB-C dataset and compare it with the state-of-the-art occlusion robust method named InterpretFR~\cite{yin2019towards}. The results are shown in Table~\ref{tab:IJBC_tatal}, where the result of InterpretFR is directly quoted from~\cite{yin2019towards}. As seen, our LCD outperforms InterpretFR with an improvement up to 2.2\% of TAR when FAR = 0.001, which demonstrates the effectiveness of our locality-aware channel-wise dropout and spatial attention module. Besides, we compare with the image augmentation method Cutout~\cite{devries2017improved}, which randomly puts black box on faces to simulate occlusions. Attributed to simulating more realistic occlusion by LCD, our method significantly surpasses Cutout with 15.1\% improvement in terms of TAR when FAR = 0.0001. Since DropBlock~\cite{ghiasi2018dropblock} follows the similar spirit by randomly zeroing out continuous activations of all channels to enhance the feature representations, we conduct further comparison with DropBlock and our method also significantly outperforms it, demonstrating the superiority of proposing LCD for occlusion synthesis. Detailed analysis between our method and DropBlock are illustrated in Section~\ref{sec:discussion}.

To further verify the effectiveness of our LCD for tackling face recognition under occlusions, we make more challenging experiments on the IJB-C occlusion subset, which only consists of occluded faces. Similar conclusion can be achieved that our LCD outperforms InterpretFR with an improvement up to 2.0\% in terms of TAR when FAR = 0.001, demonstrating the superiority of our method again. It is worth mentioning that our LCD significantly surpasses Cutout and DropBlock with improvements up to 28.4\% and 25.3\% in terms of TAR when FAR = 0.0001, respectively. Our LCD can simulate more realistic occlusions than both Cutout and DropBlock, which markedly improves the robustness to occlusions for face recognition. Furthermore, comparing to the weighted channel-wise dropout (WCD), our method achieves an improvement up to 3.3\% of TAR when FAR = 0.001, demonstrating that our method is more superior in dealing with occluded face recognition. Detailed analysis between our method and WCD are illustrated in Section~\ref{sec:discussion}.

\subsubsection{Evaluations on the LFW benchmark}

Table~\ref{tab:LFW} summarizes the accuracy results on the LFW dataset. As seen, our LCD performs better than the occlusion discarding method PDSN~\cite{song2019occlusion}. Furthermore, when comparing with stronger competitor CurricularFace~\cite{huang2020curricularface} which utilizes a larger backbone of ResNet100 and much more training images, our LCD still achieves a comparable result of 99.78\%. It is worth noting that the LFW is a general face recognition benchmark which mainly consists of non-occluded faces. Therefore, these results indicates that our method also generalizes well under non-occluded scenarios.

\begin{table}[t]
\caption{Verification performance (\%) of various methods on LFW dataset. \#Image is the number of images used for training.}
\label{tab:LFW}	
\centering
\begin{tabular}{ p{3cm}p{1cm}<{\centering} p{1cm}<{\centering}}
\hline
\multicolumn{1}{c }{Methods} & \#Image       & Accuracy \\ \hline
DeepID~\cite{sun2014deep} &0.1M &99.47\\
Deep Face~\cite{taigman2014deepface} &4.4M& 97.35\\
VGG Face~\cite{parkhi2015deep} &2.6M &98.95\\
FaceNet~\cite{schroff2015facenet} &200M &99.63\\
Baidu~\cite{liu2015targeting} &1.3M&99.13 \\
Center Loss~\cite{wen2016discriminative}&0.7M&99.28\\
Range Loss~\cite{zhang2017range}&5M & 99.52\\
Marginal Loss~\cite{deng2017marginal} &3.8M &99.48\\
SphereFace~\cite{liu2017sphereface} &0.5M&99.42\\
SpherFace+~\cite{liu2018learning} &0.5M&99.47\\
PDSN~\cite{song2019occlusion}&0.5M& 99.20\\
CosFace~\cite{wang2018cosface}&5M&99.73\\
ArcFace, R100~\cite{deng2019arcface} &5.8M& 99.83 \\
MV-Arc-Softmax~\cite{wang2020mis} &3.28M&99.78\\
CurricularFace, R100~\cite{huang2020curricularface}&5.8M&99.80\\\hline
LCD (ours) &3.8M& 99.78    \\ \hline\end{tabular}
\end{table}

\subsubsection{Evaluations on the MegaFace benchmark}
Finally, we evaluate our method on the MegaFace which is a more challenging benchmark for general scenarios. Table~\ref{tab:megaface} shows the rank-1 accuracies of recent methods on this benchmark. By leveraging only 3.8 million face images for training, our method with a light backbone of ResNet-50 surpasses both ArcFace ~\cite{deng2019arcface} and CurricularFace~\cite{huang2020curricularface} which use ResNet-100 as a backbone and are trained with 5.8 million face images. Attributed to the LCD which encourages the network to learn more comprehensive features from faces, our method not only improves the robustness to occlusions but also enhances the feature representation learning for general face recognition.

\begin{table}[t]
\caption{Rank-1 identification accuracy (\%) on MegaFace Challenge 1. \#Image is the number of images used for training.}
\label{tab:megaface}	
\centering
\begin{tabular}{ p{3cm}p{1cm}<{\centering} p{1cm}<{\centering}}
\hline
\multicolumn{1}{c }{Methods}         & \#Image & MF1\\ \hline
RegularFace~\cite{zhao2019regularface} &3.1M&75.61\\
UniformFace~\cite{duan2019uniformface} &3.8M&79.98\\
CosFace~\cite{wang2018cosface} &5.0M&82.72\\
ArcFace, R100~\cite{deng2019arcface} &5.8M&81.03 \\
PFE~\cite{shi2019probabilistic} &4.4M&78.95\\
CurricularFace, R100~\cite{huang2020curricularface} &5.8M&81.26\\\hline
LCD (ours) &3.8M&83.57\\ \hline\end{tabular}
\end{table}

\section{Conclusion}
\label{sec:conclusion}
Different from previous methods which augment face images with synthesized occlusions, we propose a novel method to better simulate realistic occlusions by dropping a group of activations in intermediate features. We first employ a spatial regularization to encourage each feature channel to respond to different face regions. In this way, the activations affected by the partial occlusion are more likely to be located in a single feature channel. Then, the locality-aware channel-wise dropout is proposed to simulate the occlusion by dropping out the entire feature channel. In addition, we design an auxiliary spatial attention module to reweight the feature channels, which can further emphasize the contributions of non-occluded regions. By directly simulating the influence of arbitrary occlusion on intermediate features, the proposed method improves the robustness against occlusion by encouraging the neural network to capture more discriminative information from the non-occluded face regions. Extensive experiments on various benchmarks have shown that the proposed method is a practical and effective approach which outperforms state-of-the-art methods with a remarkable improvement. From the practice in this work, we can conclude that the well-known dropout strategy is not only effective for improving the generalizability but also good for achieving occlusion robustness after simple modification. Our work also shows that the simulation of occlusion in feature-level rather than image-level can be a good direction to further study.

% or


% @todo_appendices
% \appendices

% % you can choose not to have a title for an appendix
% % if you want by leaving the argument blank
% \section{}


\ifCLASSOPTIONcaptionsoff
  \newpage
\fi



\bibliographystyle{IEEEtran}
\bibliography{reference}
\begin{IEEEbiography}[{\includegraphics[width=1in,height=1.25in,clip,keepaspectratio]{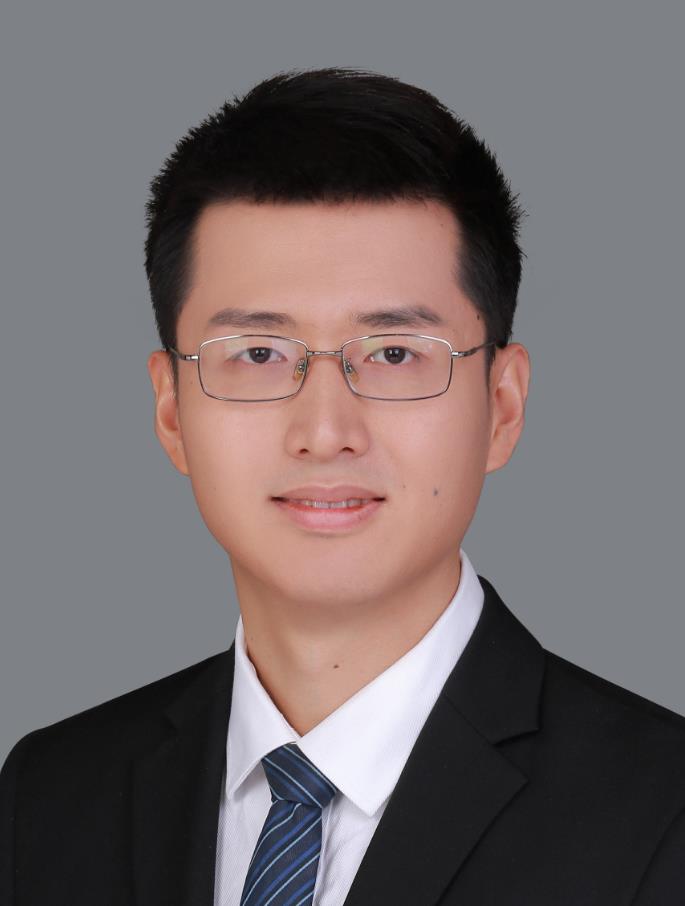}}]{Mingjie He} received the M.S. degree from the University of Science and Technology of China, Hefei, China, in 2014. Currently, he is a Ph.D. candidate at the University of Chinese Academy of Sciences and an engineer with the Institute of Computing Technology, Chinese Academy of Sciences (CAS). His research interests cover computer vision and machine learning.

\end{IEEEbiography}

% \vfill
\begin{IEEEbiography}[{\includegraphics[width=1in,height=1.25in,clip,keepaspectratio]{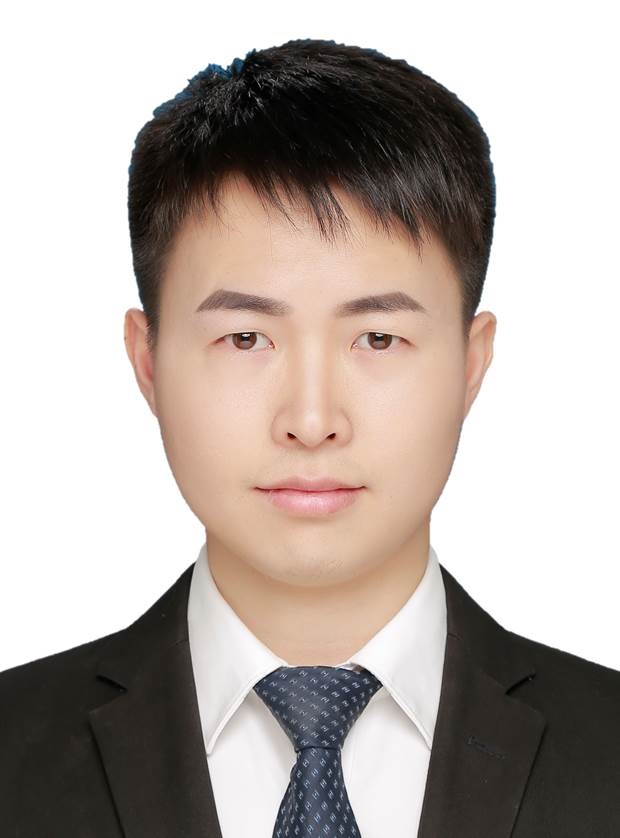}}]{Jie Zhang} is an associate professor with the Institute of Computing Technology, Chinese Academy of Sciences (CAS). He received the Ph.D. degree from the University of Chinese Academy of Sciences, Beijing, China. His research interests cover computer vision, pattern recognition, machine learning, particularly include face recognition, image segmentation, weakly/semi-supervised learning, domain generalization.

\end{IEEEbiography}

% \vfill
\begin{IEEEbiography}[{\includegraphics[width=1in,height=1.25in,clip,keepaspectratio]{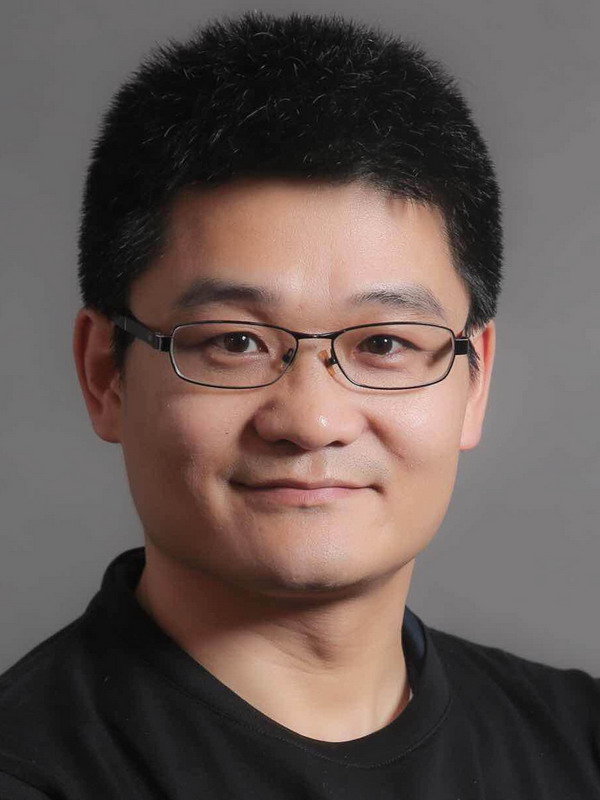}}]{Shiguang Shan} received Ph.D. degree in computer science from the Institute of Computing Technology (ICT), Chinese Academy of Sciences (CAS), Beijing, China, in 2004. He has been a full Professor of this institute since 2010 and now the deputy director of CAS Key Lab of Intelligent Information Processing. His research interests cover computer vision, pattern recognition, and machine learning. He has published more than 300 papers, with totally more than 15,000 Google scholar citations. He has served as Area Chair (or Senior PC) for many international conferences including ICCV11, ICPR12/14/20, ACCV12/16/18, FG13/18, ICASSP14, BTAS18, AAAI20/21, IJCAI21, and CVPR19/20/21. And he was/is Associate Editors of several journals including IEEE T-IP, Neurocomputing, CVIU, and PRL. He was a recipient of the China's State Natural Science Award in 2015, and the China's State S\&T Progress Award in 2005 for his research work.
\end{IEEEbiography}

% \vfill
\begin{IEEEbiography}[{\includegraphics[width=1in,height=1.25in,clip,keepaspectratio]{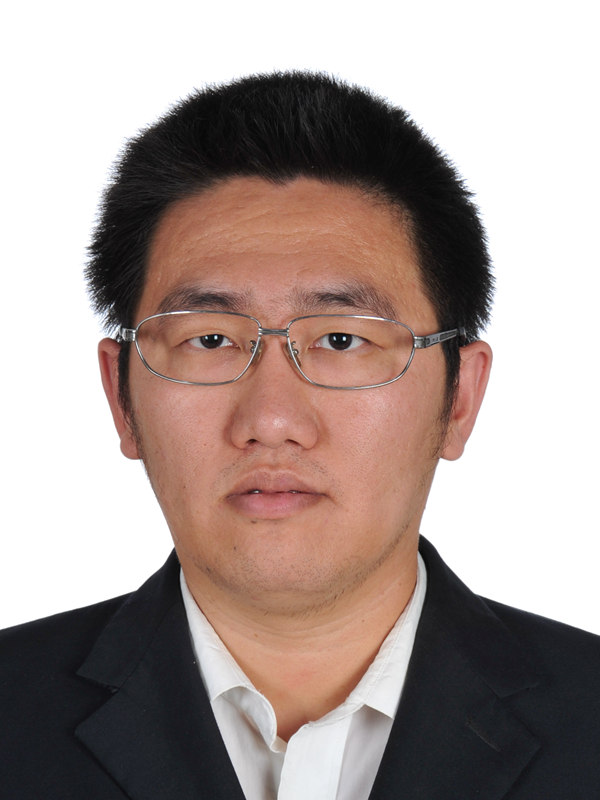}}]{Xiao Liu} is a researcher with Tomorrow Advancing Life Education Group (TAL). He received the PhD degree in computer science from Zhejiang University in 2015 and then worked for Baidu from 2015 to 2019. His research interests is the applied AI such as intelligent multimedia processing, computer vision and learning system. His research results have expounded in 40+ publications at journals and conferences such as  IEEE T-IP, T-NNLS, CVPR, ICCV, ECCV, AAAI and MM. As a key team member, he achieved the best performance in various competitions, such as the ActivityNet challenges, NTIRE super resolution challenge, EmotioNet facial expression recognition challenge, etc.
\end{IEEEbiography}

% \vfill
\begin{IEEEbiography}[{\includegraphics[width=1in,height=1.25in,clip,keepaspectratio]{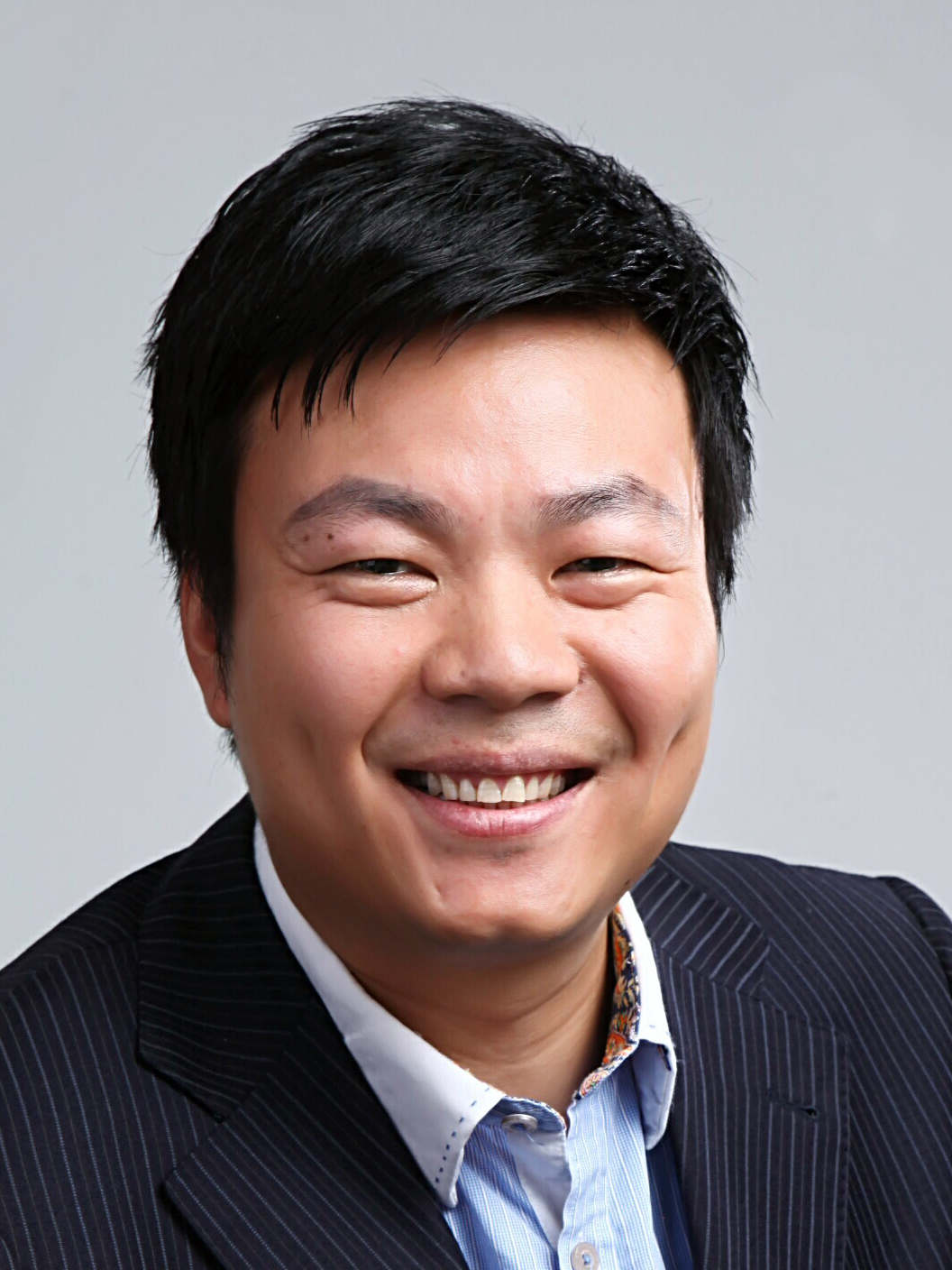}}]{Zhongqin Wu} is a scientist in the Tomorrow Advancing Life Education Group (TAL) and in charge of the AI laboratory. He received the master degree in computer science from Fudan University, China. He previously worked for Baidu and leaded the Department of Computer Vision and Augmented Reality Laboratory.
\end{IEEEbiography}

% \vfill
\begin{IEEEbiography}[{\includegraphics[width=1in,height=1.25in,clip,keepaspectratio]{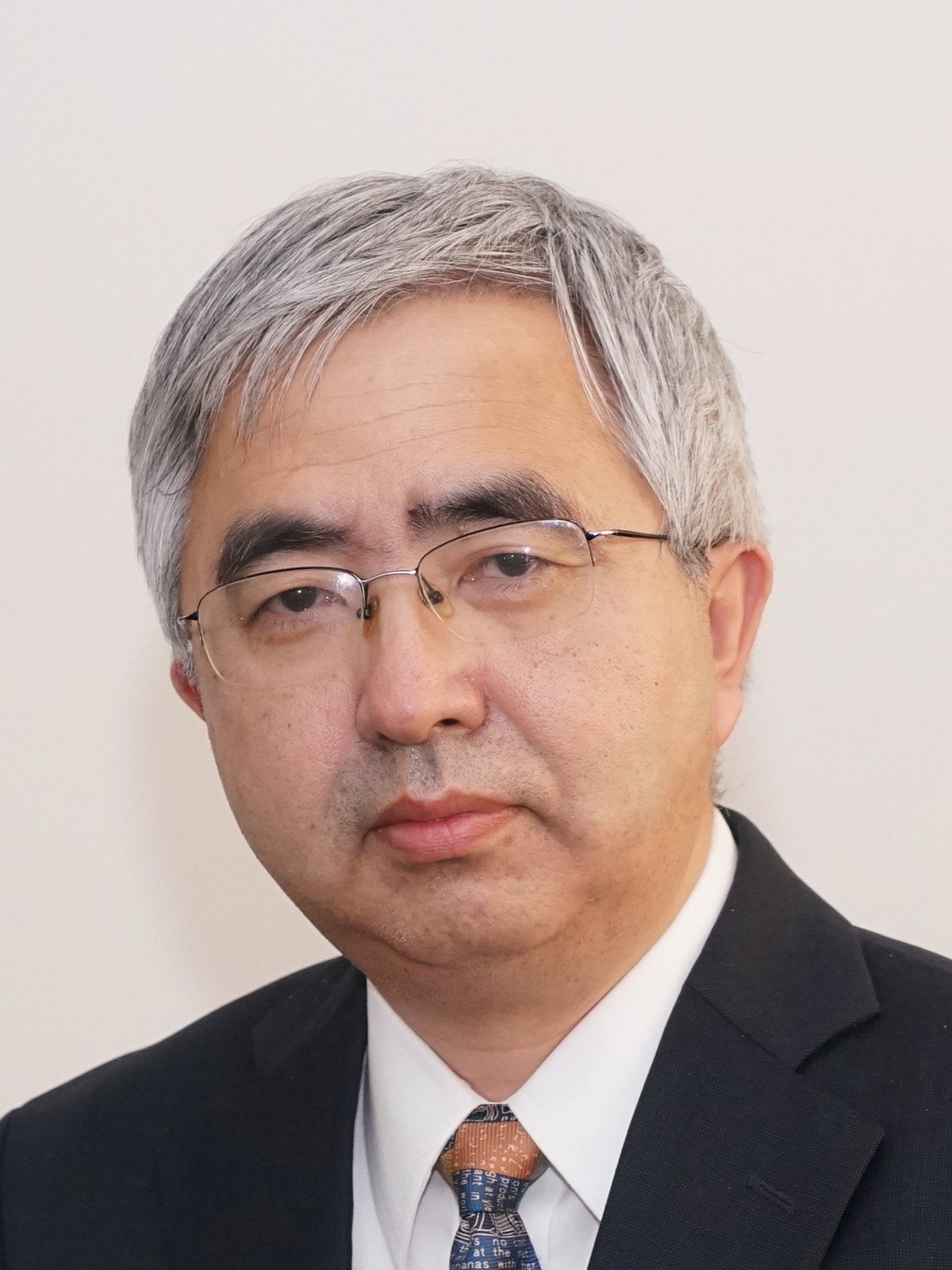}}]{Xilin Chen} is a professor with the Institute of Computing Technology, Chinese Academy of Sciences (CAS). He has authored one book and more than 200 papers in refereed journals and proceedings in the areas of computer vision, pattern recognition, image processing, and multimodal interfaces. He was a recipient of several awards, including the China's State Natural Science Award in 2015, the China's State S\&T Progress Award in 2000, 2003, 2005, and 2012 for his research work. He is currently an associate editor of the IEEE Transactions on Multimedia, and Journal of Visual Communication and Image Representation, a leading editor of the Journal of Computer Science and Technology, and an associate editor-in-chief of the Chinese Journal of Computers, and Chinese Journal of Pattern Recognition and Artificial Intelligence. He served as an Organizing Committee member for many conferences, including general co-chair of FG13/FG18, Local chair of ICME07, ACM MM09, and ICIP17, and Finance chair of ISCAS13. He is/was an area chair of CVPR 2017/2019/2020, and ICCV 2019. He is a fellow of the ACM, IEEE, IAPR, and CCF.
\end{IEEEbiography}

\end{document}